\documentclass[11pt]{article}
\usepackage[final]{acl}

\usepackage{times}
\usepackage{latexsym}

\usepackage[T1]{fontenc}
\usepackage[utf8]{inputenc}
\usepackage{microtype}
\usepackage{inconsolata}
\usepackage{booktabs}
\usepackage{graphicx}
\usepackage{amsmath}
\usepackage{xcolor}
\usepackage{enumitem}

\usepackage{tikz}
\usetikzlibrary{positioning, arrows.meta}
\usepackage{fontawesome5}
\usepackage{xurl}
\hypersetup{
  pdftitle={RA-DPO: Reliability-Aware Preference Optimization and Selective Prediction for Sexism Detection},
  pdfauthor={Hadi Mohammadi, Shihan Wang, Masoume M. Raeissi, Anastasia Giachanou}
}

\title{RA-DPO: Reliability-Aware Preference Optimization and Selective Prediction for Sexism Detection}

\author{
  \textbf{Hadi Mohammadi}\textsuperscript{1}\thanks{Corresponding author: \texttt{h.mohammadi@uu.nl}}\quad
  \textbf{Shihan Wang}\textsuperscript{2}\quad
  \textbf{Masoume M. Raeissi}\textsuperscript{1,3}\quad
  \textbf{Anastasia Giachanou}\textsuperscript{1}\\[3pt]
  \textsuperscript{1}Department of Methodology, Statistics \& Data Science, Utrecht University, Netherlands\\
  \textsuperscript{2}Department of Information and Computing Sciences, Utrecht University, Netherlands\\
  \textsuperscript{3}Field Technology Innovations \& Robotics, Wageningen University \& Research, Netherlands
}

\begin{document}
\maketitle

\begin{center}
  \footnotesize
  \textcolor{red!70!black}{\faExclamationTriangle\enspace This paper contains an example of sexist language.}
\end{center}

\begin{abstract}
Sexism detection is subjective, yet most systems reduce multi-annotator labels to a single majority label, discarding annotator agreement and model uncertainty. We propose RA-DPO (Reliability-Aware Direct Preference Optimization), a framework built on standard DPO that combines annotator agreement, model confidence and token-level uncertainty into one reliability score: agreement and token-level uncertainty select training pairs, agreement up-weights them, and the full score decides when the model should abstain. We evaluate on EXIST 2023 and EDOS, fine-tuning gpt-4o and two open-weight 3B models. The score matters most at inference: on gpt-4o, selective prediction lifts RA-DPO to 96.2\% accuracy at 50\% coverage with true agreement and 88.7\% with agreement predicted from text, against 87.9\% for its confidence alone, and abstention improves every variant on both corpora. On the open-weight models, however, confidence alone ranks RA-DPO and Standard DPO better than the full score. On the training side, fewer pairs suffice (30\% matches full-data DPO on gpt-4o), but size-matched random subsets (50\% on gpt-4o, 30\% on the 3B models) do as well as reliability-ranked ones.
\end{abstract}
\centerline{\small\faGithub\enspace\href{https://github.com/mohammadi-hadi/RA-DPO}{\texttt{github.com/mohammadi-hadi/RA-DPO}}}
\medskip

\section{Introduction}

Annotator disagreement is a central challenge in subjective NLP tasks, and sexism detection is a particularly clear example. Inter-annotator agreement on sexism labels is consistently lower than in less subjective classification tasks, and recent shared tasks report persistent disagreement among annotators \citep{kirk2023edos, plaza2023exist}. This disagreement reflects both content properties, such as subtlety and ambiguity, and annotators' attitudes and beliefs \citep{jiang2024re, mohammadi2025assessing}. Prior work has treated disagreement as a useful signal that captures different human perspectives \citep{plank2022problem, davani2022dealing, wu-etal-2023-dont, pandya2024exploring}. In content moderation, the practical question is then when a model should pass a post to human review rather than assign a label itself.


\begin{figure}[t]
\centering
\resizebox{0.90\columnwidth}{!}{%
\begin{tikzpicture}[
  font           = \scriptsize,
  box/.style     = {draw, rounded corners=2pt, align=center,
                    inner sep=3pt, line width=0.4pt},
  example/.style = {box, fill=black!4, minimum width=74mm, minimum height=13mm,
                    line width=0.5pt},
  signal/.style  = {box, fill=blue!8, minimum width=22mm, minimum height=8mm},
  rx/.style      = {box, fill=orange!18, minimum width=74mm, minimum height=10mm,
                    line width=0.7pt},
  train/.style   = {box, fill=green!14, minimum width=35mm, minimum height=13mm},
  infer/.style   = {box, fill=violet!14, minimum width=35mm, minimum height=13mm},
  arr/.style     = {-{Stealth[length=3.5pt]}, line width=0.4pt, draw=black!75},
]

\node[example] (in) {%
  Tweet: ``Women in politics are too emotional to lead''\\[1pt]
  Labels: 5 YES, 1 NO \quad Model output: YES
};

\node[signal, below=2mm of in, xshift=-25mm] (c)
  {$\mathrm{conf} = 0.87$};

\node[signal, below=2mm of in] (a)
  {$\mathrm{agree} = 5/6$};

\node[signal, below=2mm of in, xshift=25mm] (s)
  {$1 - \mathrm{sig} = 0.75$};

\node[rx, below=2mm of a] (rx) {%
  $R_{\mathrm{train}}(x) = 0.8 \cdot \mathrm{agree} + 0.2 \cdot (1 - \mathrm{sig}) = 0.82$\\[1pt]
  $R(x) = \alpha \cdot \mathrm{conf}
  + \beta \cdot \mathrm{agree}
  + \gamma \cdot (1 - \mathrm{sig}) = 0.83$
};

\node[train, below=2mm of rx, xshift=-19mm] (tr) {%
  \textbf{Training}\\
  rank pairs by $R_{\mathrm{train}}(x)$;\\
  Smart: DPO on top-ranked pairs;\\
  RA-DPO: 2$\times$ if agree $\geq 5/6$
};

\node[infer, below=2mm of rx, xshift=19mm] (inf) {%
  \textbf{Inference}\\
  answer if $R(x) \geq \tau$;\\
  else abstain
};

\draw[arr] (in.south) -- (c.north);
\draw[arr] (in.south) -- (a.north);
\draw[arr] (in.south) -- (s.north);

\draw[arr] (c.south) -- (rx.north);
\draw[arr] (a.south) -- (rx.north);
\draw[arr] (s.south) -- (rx.north);

\draw[arr] (rx.south) -- (tr.north);
\draw[arr] (rx.south) -- (inf.north);

\end{tikzpicture}%
}

\caption{RA-DPO framework with synthetic values.}
\label{fig:pipeline}
\end{figure}

The EXIST 2023 shared task on sEXism Identification in Social neTworks \citep{plaza2023exist} preserves this signal explicitly: each post is labelled by six annotators, and disagreement is retained rather than collapsed into a single gold label. However, most systems still train on a single aggregated label and treat all examples uniformly, discarding two useful signals about example reliability. The first is annotator agreement, which reflects how reliable the human labels are for an example. The second is model confidence, both at the example level and at the token level. Existing reliability-aware methods typically use only one of these signals. Soft-label methods use annotator agreement at training only \citep{wu-etal-2023-dont}, while token-level preference methods such as TIS-DPO \citep{liu2025tisdpo} and ConfPO \citep{yoon2025confpo} use model-derived token probabilities at training only. This leaves a gap for a unified reliability score that combines external and internal signals and can be used consistently for training-time selection and inference-time abstention.

We propose RA-DPO, a framework built on Direct Preference Optimization \citep[DPO;][]{rafailov2023dpo} that leaves the preference objective unchanged and uses reliability signals during both training and inference. We build on DPO because a preference pair is a natural unit for encoding how far each label can be trusted: reliability signals decide which pairs enter training and how often, and the full score, with the deployed model's confidence added, decides when the model abstains. For each example, $R(x)$ combines three factors: model prediction confidence, annotator agreement, and token-level uncertainty. During training, the part of the score available before fine-tuning (agreement and token-level uncertainty) ranks preference pairs: the Smart-$k\%$ variants fine-tune with DPO on a subset drawn from the highest-ranked pairs, and the RA-DPO variant keeps every pair but gives those with high agreement a second copy. At inference, $R(x)$ controls abstention: the model predicts only when $R(x)$ exceeds a learned threshold, trading coverage for accuracy (Figure~\ref{fig:pipeline}).

We benchmark nine OpenAI models under five prompting strategies, fine-tune gpt-4o and two open-weight 3B backbones (Llama and Qwen), and introduce RA-DPO as a reliability-aware training-and-inference framework, evaluated on two sexism corpora that publish per-annotator labels, EXIST 2023 and EDOS (Explainable Detection of Online Sexism). Our strongest result is at inference: on gpt-4o with true annotator agreement, abstaining on posts with a low $R(x)$ lifts RA-DPO to 96.2\% accuracy at 50\% coverage, against 87.9\% when abstaining on confidence alone (Section~\ref{sec:results}).

\section{Related Work}

\paragraph{Sexism detection.}
Transformer encoders such as BERT \citep{devlin2019bert} and RoBERTa \citep{liu2019roberta} established strong baselines for sexism detection \citep{kirk2023edos, mohammadi2023robust}. Related work targets the more specific phenomenon of online misogyny on expert-annotated corpora \citep{guest2021misogyny}, which complements the larger multi-annotator datasets we build on. Explainable approaches further combine model predictions with token-level rationales to make outputs interpretable \citep{mohammadi2024transparent, mohammadi2026letmeexplain}. More recent work uses large language models (LLMs) with supervised fine-tuning (SFT) and reinforcement learning from human feedback to improve both accuracy and explanation \citep{samani2025rlhf}. These approaches optimise for a single aggregated label and inherit the same issue: many annotator disagreements reflect valid differences in interpretation rather than errors, and that signal is lost when the labels are collapsed. Our work targets this gap by keeping annotator agreement as a main signal at training and inference.

\paragraph{DPO and preference-data quality.}
In DPO \citep{rafailov2023dpo}, a model learns to prefer chosen responses over rejected ones without a separate reward model. Later work improves it either by modifying the loss, as in KTO \citep{ethayarajh2024kto}, or by improving the preference data, as in Filtered DPO \citep{morimura2024fdpo}, which shows DPO is sensitive to pair quality. We follow the data branch but filter on an external signal, annotator agreement via $R_{\mathrm{train}}(x)$ (Section~\ref{sec:methodology}), rather than on within-model preference quality. VPO \citep{cho2026vpo} also uses annotator vote counts, as a soft target inside the loss, and large-scale comparisons of instruction-tuning data selection find scored subsets rarely improve on random ones \citep{xia2025rethinking}, which our matched-budget controls confirm (Section~\ref{sec:results}). RA-DPO leaves the DPO objective unchanged; its contribution is the reliability score itself, which selects pairs and decides when to abstain.

\paragraph{Token-level DPO variants.}
Some recent methods refine DPO at the token level. TIS-DPO \citep{liu2025tisdpo} weights tokens by their estimated importance during preference training. ConfPO \citep{yoon2025confpo} picks critical tokens using the model's own confidence and updates only those tokens. Our $R(x)$ borrows the token-probability intuition from these methods but applies it at the example level: the token-uncertainty factor is one of three signals in $R(x)$, alongside example-level confidence and annotator agreement. A second difference is that model confidence enters only at inference, where $R(x)$ gates abstention; training ranks pairs by agreement and token-level uncertainty ($R_{\mathrm{train}}$), whereas TIS-DPO and ConfPO use their token signals inside the training loss.

\paragraph{Disagreement as signal.}
A growing line of work treats annotator disagreement as an informative signal rather than noise \citep{uma2021learning, plank2022problem, davani2022dealing, mohammadi2026perspectivist}, framed by \citet{rottger2022paradigms} as a choice between descriptive and prescriptive annotation. \citet{fornaciari2021soft} train on soft labels derived from disagreement under a multi-task objective, and \citet{wu-etal-2023-dont} build soft labels from annotator confidence, secondary labels and disagreement for single-label classification of COVID-19 vaccine-stance tweets and COVID-19 disinformation debunks and report gains over hard-label training on both datasets. It is close to our training side, as it also uses per-example agreement as a training signal. We differ in two ways: we combine agreement with the model's own confidence and token-level uncertainty into one abstention score $R(x)$, whose agreement and token-uncertainty part also ranks training pairs, and on our evidence the inference side is where it matters most. \citet{fleisig2023majority} and \citet{pandya2024exploring} further show majority-vote aggregation can marginalise minority annotator views, motivating the predicted-agreement variant of Section~\ref{sec:results}.

\paragraph{Selective prediction.}
A complementary line of work treats abstention as a way to trade coverage for accuracy. \citet{geifman2017selective} formalised the coverage-risk trade-off for deep classifiers, and \citet{kamath2020selective} applied it to question answering under domain shift, learning a calibrator that decides when the model should abstain. \citet{varshney2022investigating} compared selective-prediction methods across 17 NLP datasets and reported that the model's own softmax confidence (MaxProb) is a strong baseline that is difficult to exceed, which our results confirm: on the open-weight backbones confidence alone outranks the combined score for RA-DPO and Standard DPO, and on gpt-4o SFT does well at $50\%$ coverage because $R(x)$ gives it a large confidence weight. RA-DPO follows this line of work, with two differences: $R(x)$ combines internal model confidence with the external signal of annotator agreement, and the agreement and token-uncertainty signals that rank training pairs also enter the abstention score, so selection and abstention draw on the same annotator signal.

\paragraph{Prompting and in-context learning.}
Prompt design strongly affects LLM performance on classification tasks. Chain-of-Thought (CoT) prompting \citep{wei2022cot} encourages step-by-step reasoning, and persona-based prompting asks the model to act as a domain expert. \citet{muscato2026seeingallsides} evaluate in-context learning for modelling human disagreement and find that prompt design and example selection affect performance on subjective tasks. Our prompt-strategy study (Section~\ref{sec:results}) addresses the same question, but we pair it with reliability-aware fine-tuning and inference; prompting sets the baseline for the rest of the paper.

\section{Methodology}
\label{sec:methodology}

\paragraph{Reliability score.}
RA-DPO extends DPO with a three-factor score computed as:
\begin{equation}
\begin{split}
R(x) = {}& \alpha \cdot \mathrm{conf}(x) + \beta \cdot \mathrm{agree}(x)\\
&+ \gamma \cdot (1 - \mathrm{sig}(x))
\end{split}
\label{eq:reliability}
\end{equation}
where:

\begin{itemize}
\item $\mathrm{conf}(x) \in [0,1]$ is the model's confidence, the softmax probability of the predicted token.
\item $\mathrm{agree}(x) \in \{0.5,\, 0.667,\, 0.833,\, 1.0\}$ is the annotator agreement, the fraction of annotators who chose the majority label.
\item $\mathrm{sig}(x) = \mathrm{mean}\bigl(\sigma(k \cdot (T - p_i))\bigr) \in [0, 1]$ is a token-level uncertainty score, where $p_i$ are token probabilities from a fixed scoring model, computed once per post before fine-tuning and shared by all variants, $T$ is the mean token probability, $k = 10$ is a steepness parameter, and $\sigma$ is the sigmoid. Higher $\mathrm{sig}(x)$ means more uncertain tokens, so $1 - \mathrm{sig}(x)$ is higher for examples with fewer uncertain tokens.
\end{itemize}

We learn $\alpha, \beta, \gamma$ by 5-fold stratified cross-validation, so every reported weight is out-of-fold. On each fold, a logistic regression takes $(\mathrm{conf}, \mathrm{agree}, 1 - \mathrm{sig})$ as features and predicts whether the model's answer is correct; the normalised absolute coefficients give the per-fold weights, and we report their mean in Table~\ref{tab:weights}. Every test example's $R(x)$ uses weights fit on the other folds, so the coverage-accuracy numbers below are leakage-free. We normalise so that $\alpha + \beta + \gamma = 1$.

\begin{table}[t]
\centering
\small
\setlength{\tabcolsep}{4pt}
\begin{tabular}{llccc}
\toprule
Backbone & Method & $\alpha$ & $\beta$ & $\gamma$ \\
\midrule

OpenAI & Base            & 0.378 & 0.536 & 0.086 \\
        & SFT             & 0.280 & 0.682 & 0.038 \\
        & Smart-10\%        & 0.055 & 0.897 & 0.048 \\
        & Smart-30\%        & 0.084 & 0.843 & 0.073 \\
        & Smart-50\%        & 0.113 & 0.836 & 0.052 \\
        & Random-50\%       & 0.129 & 0.791 & 0.079 \\
        & Standard DPO    & 0.059 & 0.809 & 0.133 \\
        & Ambiguous-only  & 0.171 & 0.621 & 0.207 \\
        & RA-DPO          & 0.146 & 0.827 & 0.027 \\

\midrule

OpenAI  & Shared          & 0.131 & 0.793 & 0.076 \\
        & RA-DPO (pred.)  & 0.168 & 0.762 & 0.070 \\

\midrule

Qwen    & SFT             & 0.082 & 0.608 & 0.310 \\
        & Smart-10\%        & 0.230 & 0.500 & 0.269 \\
        & Smart-30\%        & 0.216 & 0.515 & 0.269 \\
        & Smart-50\%        & 0.164 & 0.591 & 0.245 \\
        & Random-50\%       & 0.142 & 0.608 & 0.250 \\
        & Standard DPO    & 0.158 & 0.591 & 0.251 \\
        & Ambiguous-only  & 0.275 & 0.487 & 0.239 \\
        & RA-DPO          & 0.147 & 0.624 & 0.229 \\

\midrule

Llama   & SFT             & 0.637 & 0.164 & 0.199 \\
        & Smart-10\%        & 0.358 & 0.335 & 0.308 \\
        & Smart-30\%        & 0.067 & 0.535 & 0.398 \\
        & Smart-50\%        & 0.066 & 0.673 & 0.261 \\
        & Random-50\%       & 0.042 & 0.646 & 0.312 \\
        & Standard DPO    & 0.240 & 0.572 & 0.189 \\
        & Ambiguous-only  & 0.086 & 0.440 & 0.475 \\
        & RA-DPO          & 0.193 & 0.668 & 0.140 \\

\bottomrule
\end{tabular}
\caption{Out-of-fold $R(x)$ weights per backbone and variant. Shared: one set pooled over gpt-4o base and five DPO variants; pred.: predicted agreement.}
\label{tab:weights}
\end{table}

\paragraph{Smart sampling.} For every training example, we build a preference pair: the majority-vote label is the chosen response, and the opposite label is the rejected one. Standard DPO trains on all 5,536 pairs with equal weight.

Instead of training DPO on all available data, we propose smart sampling that trains only on the pairs the reliability score ranks highest. At training time the model that will be fine-tuned does not yet exist, so the ranking uses the two signals available before training: \mbox{$R_{\mathrm{train}}(x) = 0.8 \cdot \mathrm{agree}(x) + 0.2 \cdot (1 - \mathrm{sig}(x))$}; the confidence term joins the full $R(x)$ (Eq.~\ref{eq:reliability}) at inference, where it comes from the deployed model. In the EXIST dataset, we rank all 5,536 pairs by $R_{\mathrm{train}}(x)$ and build six variants. Smart-30\% takes the top 1,661 pairs, and Smart-10\% the first 554 of these in training-file order, not the top 554 by $R_{\mathrm{train}}$. Smart-50\% takes the top 2,768 pairs. Random-50\% takes 2,768 pairs at random as a data-size control, and Random-30\% takes 1,661 at random, size-matched to Smart-30\%. Ambiguous-only takes the 665 pairs with agreement exactly $0.5$ (3/3 splits) as a negative control to test whether ambiguous pairs improve or degrade performance. Similar subsets were also sampled in the EDOS dataset. 

We compare these against Standard DPO (all 5,536 pairs) and RA-DPO. RA-DPO trains on the same 5,536 unique pairs but duplicates the high-reliability ones (two copies when agreement $\geq 5/6$, one otherwise), giving 8,984 training rows; this weighted duplication is what the $0.62\times$ entry in the data-efficiency column of Table~\ref{tab:finetuning} reflects. The composition analysis below shows that every Smart-$k\%$ subset consists mostly of unanimous-label pairs; because size-matched random subsets reach the same F1, the early flattening of the F1 curve reflects how quickly the task saturates rather than subset composition (F1 is macro-averaged throughout).

\paragraph{Abstention.}
At inference, we compute $R(x)$ for each test example and answer when $R(x) \geq \tau$; otherwise the system abstains. We choose $\tau$ by trying values from $0.3$ to $0.95$ and picking the one that maximises the harmonic mean of accuracy and coverage,
\begin{equation}
H(\mathrm{acc}, \mathrm{cov}) = \frac{2 \cdot \mathrm{acc} \cdot \mathrm{cov}}{\mathrm{acc} + \mathrm{cov}},
\label{eq:harmonic}
\end{equation}
so that $\tau$ balances answering most of the test set against keeping accuracy on the answered subset high. The coverage-accuracy tables that follow instead cut at fixed coverage levels, which makes variants comparable at equal answer rates; $\tau$ is the operating point a deployment would select, and Table~\ref{tab:tau_sensitivity} reports its sensitivity.

\paragraph{Datasets.}
Computing $R(x)$ requires labels from each annotator, which rules out the many corpora that release only an aggregated gold label. We evaluate on two sexism corpora that publish them, and their differing annotation protocols are what let us ask how the reliability score behaves as the agreement signal changes.

EXIST 2023 \citep{plaza2023exist} is the standard multilingual benchmark and preserves all six annotator labels per post under the ``learning with disagreement'' paradigm. We use Task 1 (binary sexism detection) on the 6,920 posts of the official training release, in English and Spanish, each labelled by six annotators as sexist (YES) or not sexist (NO), split into train (5,536), validation (692), and test (692) by stratified sampling; our scores are therefore not comparable to the shared-task leaderboard.

EDOS \citep{kirk2023edos} is English-only and larger: 20,000 posts on the official splits (train 14,000 / dev 2,000 / test 4,000), each with three annotator labels and an expert-adjudicated gold label. With three annotators, agreement has only two levels (unanimous or two-to-one; 22\% of items are disputed), against EXIST's four. The test split is also $5.8\times$ larger, so effects that fall inside the confidence intervals (CIs) on EXIST are detectable here. Because the gold label is adjudicated rather than a crowd majority, it departs from the annotator majority on 946 of 20,000 items (4.7\%); agreement and correctness are coupled more loosely here than on EXIST, which is precisely the regime the reliability score should be tested in.

\paragraph{Language models.}
We compare three backbones: OpenAI, Llama, and Qwen. OpenAI is the OpenAI API base model gpt-4o \citep{openai2024gpt4o}, used for every hosted fine-tuning variant. Llama is Llama-3.2-3B-Instruct from the Llama~3 family \citep{grattafiori2024llama3}, and Qwen is Qwen2.5-3B-Instruct \citep{qwen2024qwen25}; fixing both open-weight backbones at 3B matches their scale, so they differ in pretraining data, architecture details and post-training rather than size. The eight fine-tuned variants (SFT and all DPO variants except Random-30\%) all train on the OpenAI base with a short YES/NO moderation instruction and are evaluated with the Structured prompt. Fixing a single hosted base keeps the DPO variants directly comparable. Seven further OpenAI versions and a reasoning model set the prompt-strategy baselines in zero-shot and few-shot; the highest baseline F1 on a single language subset is gpt-4o's own $0.803$ (English, Persona, few-shot). Per-version detail is in Appendix~\ref{app:openai_variants}; full hyperparameter and compute settings are in Appendix~\ref{app:hyperparams}. The token probabilities behind $\mathrm{sig}(x)$ come from a fixed scoring model: on EXIST, the probabilities gpt-4o-mini gives its score tokens when rating each word's relevance to sexism; on EDOS, Llama-3.2-3B-Instruct's teacher-forced probabilities of the post's tokens.

\paragraph{Prompt strategies.}
Before fine-tuning we identify which prompt is most informative across models. We compare five prompts that cover the main approaches in the literature, fix the best one for all later fine-tuning experiments, and use the cross-prompt comparison itself as a baseline against which the gain from fine-tuning can be measured.

Basic gives a one-line instruction: ``Classify whether this post is sexist.'' Definition adds the EXIST taxonomy with five subcategories (ideological inequality, stereotyping, objectification, sexual violence, misogyny). Chain-of-Thought asks the model to reason step by step before answering. Persona tells the model to act as a trained EXIST 2023 annotator. Structured provides an explicit checklist of five criteria, where a post that matches any criterion is classified as sexist. We test each prompt in both zero-shot (no examples) and few-shot settings (the five examples with the highest annotator agreement).

\paragraph{Settings.}
The agreement signal comes from the annotator labels (six per post on EXIST, three on EDOS), which are not available in deployment. We therefore evaluate three settings:
\begin{itemize}
\item True agreement: uses the dataset's annotator agreement; on EXIST with gpt-4o it exceeds predicted agreement at 50\% coverage for every variant except the Ambiguous-only control, while on EDOS, whose gold label is adjudicated, predicted agreement matches or exceeds it at 50\% coverage for most variants.
\item Predicted agreement: replaces true agreement with a Twitter-pretrained XLM-RoBERTa regressor \citep{barbieri2022xlmt}, trained only on the training split to predict agreement from text (Pearson $r = 0.351$ on test; Appendix~\ref{app:hyperparams}).
\item No agreement: drops $\beta$, so $R(x)$ uses confidence and token-level uncertainty.
\end{itemize}

\section{Results}
\label{sec:results}

This section reports three results: the Structured prompt gives the strongest baseline, the top six fine-tuning variants are statistically tied at full coverage, and their accuracy gaps widen under $R(x)$-gated abstention.

\paragraph{Prompt strategy.}
Over the eight base OpenAI versions, excluding o3 \citep{openai2025o3} and the fine-tuned model, Structured has the highest mean F1 ($0.731$; standard deviation $0.031$ over the 32 runs of version, shot setting and language), ahead of Definition ($0.728$, with the lowest standard deviation, $0.028$) and Persona ($0.722$). The bare Basic instruction is weakest ($0.661$) and by far the most variable (standard deviation $0.080$), and Chain-of-Thought ($0.675$) does not reach the criterion-based prompts either, so neither omitting guidance nor adding reasoning steps helps this binary decision. We fix Structured for evaluating every fine-tuned model and for every coverage experiment. Per-strategy and per-version numbers are in Appendix~\ref{app:openai_variants} (Table~\ref{tab:prompts}, Figure~\ref{fig:heatmap}).

\paragraph{Fine-tuning.}
Table~\ref{tab:finetuning} compares all fine-tuned models on the 692-sample test set (English and Spanish, Structured prompt).

\begin{table}[ht]
\centering
\small
\setlength{\tabcolsep}{2pt}
\begin{tabular}{lrrcr}
\toprule
Method & Pairs & F1 & 95\% CI & Eff. \\
\midrule

Base            & --    & 0.724 & [0.689, 0.757] & -- \\
Ambiguous-only  & 665   & 0.653 & [0.613, 0.691] & -- \\
Smart-10\%        & 554   & 0.800 & [0.769, 0.828] & 9.74$\times$ \\
Random-50\%       & 2,768 & 0.814 & [0.783, 0.842] & 1.98$\times$ \\
Smart-50\%        & 2,768 & 0.819 & [0.790, 0.847] & 2.00$\times$ \\
SFT             & 5,536 & 0.820 & [0.790, 0.849] & 1.00$\times$ \\
Standard DPO    & 5,536 & 0.821 & [0.792, 0.850] & 1.00$\times$ \\
Smart-30\%        & 1,661 & 0.821 & [0.792, 0.850] & 3.33$\times$ \\
RA-DPO          & 5,536 & 0.826 & [0.797, 0.855] & 0.62$\times$ \\

\bottomrule
\end{tabular}
\caption{Fine-tuning results on gpt-4o. Eff.: training-pair ratio times F1 ratio relative to Standard DPO; higher is better.}
\label{tab:finetuning}
\end{table}

Three points stand out. First, the top six variants (SFT, Smart-30\%, Smart-50\%, Random-50\%, Standard, RA-DPO) are statistically tied on F1: their 95\% bootstrap CIs \citep{koehn2004statistical} all overlap and pairwise McNemar \citep{mcnemar1947} $p$-values exceed $0.27$ (minimum: $p=0.28$ for Random-50\% vs RA-DPO). Second, every DPO variant of Table~\ref{tab:finetuning} except Ambiguous-only improves over the base ($p < 10^{-4}$); Ambiguous-only is the negative control and drops below the base. Third, Smart-30\% matches Standard DPO with 70\% less data ($0.821$ vs $0.821$, $p=1.00$), a $3.33\times$ data-efficiency gain. We ran the size-matched random control on every corpus and backbone: five comparisons at an identical pair budget, none significant (McNemar $p = 0.22$--$1.00$; gpt-4o Smart-50\% vs Random-50\%: $0.819$ vs $0.814$, $p = 0.69$), with the random draw numerically ahead in two (Appendix~\ref{app:ablation}). On gpt-4o those budgets sit at the full-data ceiling and on the 3B tracks within $1$--$4$ percentage points (pp) of it, so we repeated the comparison at 554 pairs, where the agreement-based selection arms and their random controls (first two blocks of Table~\ref{tab:lowbudget}) remain $3$--$12$\,pp short of it. Selection does not help there either, and restricting to the unanimous tail is worse than sampling the pool at random on Llama ($0.512$ vs $0.543$, $p = 0.03$) and no better on Qwen ($0.528$ vs $0.533$, $p = 1.00$). The penalty comes from the distribution, not the signal: ranking by $R_{\mathrm{train}}$ within each agreement stratum, so the subset keeps the pool's agreement distribution, removes the penalty on Llama and lands statistically level with random on both backbones (Table~\ref{tab:lowbudget}). A decision-threshold audit completes the picture: a threshold tuned on validation lifts the untrained 3B bases above every 554-pair variant and above full-data DPO ($0.604$ vs $0.596$ on Llama), so low-budget gains there amount to moving the operating point, whereas every fine-tuned OpenAI variant except the Ambiguous-only control exceeds even the base's best-threshold ceiling ($0.800$--$0.826$ vs $0.764$; Tables~\ref{tab:finetuning} and~\ref{tab:threshold}). We therefore read the training-side result as data efficiency, since on \mbox{gpt-4o} a few thousand pairs saturate this task, rather than as evidence that reliability ranks pairs better than chance. RA-DPO also closes the base's YES/NO imbalance: base recall is $0.89$ for NO but $0.55$ for YES, against $0.84$ and $0.81$ for RA-DPO.

\paragraph{Reliability-aware inference.}
The main benefit of RA-DPO appears when we apply $R(x)$ at inference. Table~\ref{tab:coverage} reports accuracy at matched coverage in the three agreement settings of Section~\ref{sec:methodology}.

\begin{table}[t]
\centering
\small
\setlength{\tabcolsep}{4pt}
\begin{tabular}{llccc}
\toprule
Method & Setting & @100 & @60 & @50 \\
\midrule

Base         & True      & 0.740 & 0.834 & 0.879 \\
             & Pred.     & 0.740 & 0.824 & 0.835 \\
             & No-agr.   & 0.740 & 0.793 & 0.775 \\

\midrule

SFT          & True      & 0.822 & 0.940 & 0.965 \\
Smart-10\%     & True      & 0.802 & 0.911 & 0.925 \\

Smart-30\%     & True      & 0.824 & 0.937 & 0.945 \\
             & Pred.     & 0.824 & 0.855 & 0.853 \\

Smart-50\%     & True      & 0.821 & 0.937 & 0.939 \\
Random-50\%    & True      & 0.816 & 0.935 & 0.945 \\

Standard DPO & True      & 0.825 & 0.925 & 0.922 \\
             & Pred.     & 0.825 & 0.858 & 0.867 \\

Ambiguous-only    & True      & 0.697 & 0.764 & 0.775 \\

RA-DPO       & True      & 0.828 & 0.942 & 0.962 \\
             & Pred.     & 0.828 & 0.872 & 0.887 \\
             & No-agr.   & 0.828 & 0.848 & 0.853 \\

\bottomrule
\end{tabular}
\caption{Accuracy at matched coverage for gpt-4o on EXIST under three agreement settings (True/Pred./No-agr.).}
\label{tab:coverage}
\end{table}
\begin{figure*}[t]
\centering
\includegraphics[width=\textwidth]{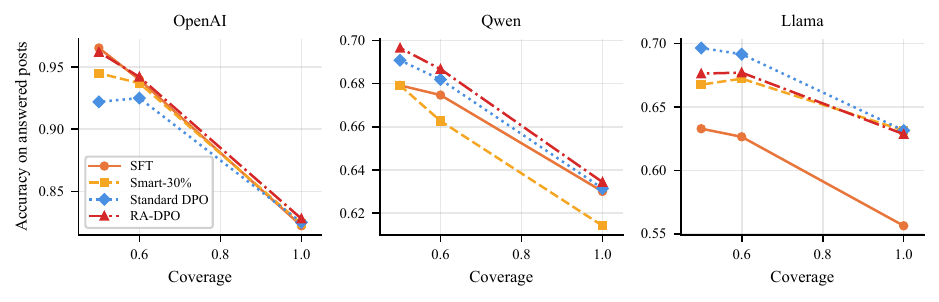}
\caption{Coverage vs.\ accuracy on EXIST across the three backbones (true-agreement setting; y-axes differ).}
\label{fig:coverage}
\end{figure*}

At 50\% coverage SFT leads at 96.5\% with RA-DPO at 96.2\%, a gap of one item on the 346 answered; SFT does well because its larger $\alpha$ weight (0.280 vs 0.146) reflects a confidence signal that ranks its own errors well (confidence alone reaches $0.951$ at 50\% coverage against $0.879$ for RA-DPO; Table~\ref{tab:rx_ablation}), itself a strong abstention cue \citep{varshney2022investigating}. For RA-DPO this is $+13.4$\,pp over its full-coverage 82.8\%. In the deployable predicted-agreement setting it reaches 88.7\%, $+3.5$\,pp above the no-agreement floor of 85.3\%, with SFT again close (89.0\%), while SFT ranked by its own confidence alone reaches 95.1\% without any annotator signal; per-variant rows for both regimes are in Appendix~\ref{app:ablation} (Table~\ref{tab:coverage-complete}). RA-DPO's lead over Standard DPO survives the move from true to predicted agreement ($+4.0$ and $+2.0$\,pp), and with one weight set shared across the base and five DPO variants (Table~\ref{tab:weights}) or uniform weights it is again the highest DPO variant at 50\% coverage (0.965 and 0.957), so its lead over the other DPO variants does not depend on how $R(x)$'s weights are set.

RA-DPO is significantly better than the base model (McNemar's paired test, $p = 1.4 \times 10^{-8}$); the gaps among the top variants widen only under abstention ($0.816$--$0.828$ at full coverage, $0.922$--$0.965$ at 50\%). Language models are often poorly calibrated \citep{jiang2021calibration}, and RA-DPO is the best calibrated of the three: the gap between mean confidence and F1, both taken over the 692 test items in the released per-instance arrays, is $0.24$ for the base ($0.967$ vs $0.724$), $0.15$ for SFT, and $0.10$ for RA-DPO ($0.925$ vs $0.826$).

Figure~\ref{fig:coverage} shows the EXIST coverage-accuracy curves for the four main fine-tuned variants on each backbone (OpenAI, Qwen, Llama). The pattern is the same on all three: accuracy rises from full coverage to 60\% on every curve, and RA-DPO is first or second of the four curves at 50\% coverage.

\paragraph{Component ablation.}
Ranking test items by each component of $R(x)$ alone, by each pair, and by the full score isolates what drives these gains (Table~\ref{tab:rx_ablation}; protocol in Appendix~\ref{app:ablation}, leave-one-out importance in Figure~\ref{fig:loo}). On gpt-4o, agreement carries most of the signal: for RA-DPO at 50\% coverage, the full score reaches $0.962$ against $0.949$ for agreement alone and $0.879$ for confidence alone (the MaxProb baseline evaluated by \citealt{varshney2022investigating}); leave-one-out removal of agreement, confidence, and token-level uncertainty costs $11.0$, $1.4$, and $0.3$\,pp respectively. On gpt-4o, confidence is the stronger single signal for the base model and SFT, and every DPO variant leans on agreement. Training-side selection shows the same reliance on agreement: the $R_{\mathrm{train}}(x)$ top-30\% consists entirely of unanimous pairs, so any agreement-only selection must overlap Smart-30\% by at least $91\%$, while uncertainty-only selection overlaps at chance level (Table~\ref{tab:selection-overlap}). On the four open-weight tracks the combination does not hold up: at 50\% coverage confidence alone gives higher accuracy than the full $R(x)$ for RA-DPO and Standard DPO on every track (RA-DPO: $0.783$ vs $0.676$ on EXIST-Llama, $0.798$ vs $0.697$ on EXIST-Qwen, $0.912$ vs $0.835$ on EDOS-Llama, $0.939$ vs $0.920$ on EDOS-Qwen).

\begin{table}[t]
\centering
\footnotesize
\setlength{\tabcolsep}{3pt}
\begin{tabular}{lccc}
\toprule
Scoring rule & @60 & @50 & AUARC \\
\midrule
\multicolumn{4}{l}{gpt-4o (base), acc@100 $=0.740$} \\
Conf-only (MaxProb) & 0.860 & 0.870 & 0.819 \\
Agree-only & 0.821 & 0.835 & 0.790 \\
$(1-\mathrm{sig})$-only & 0.706 & 0.705 & 0.674 \\
Conf $+$ Agree & 0.836 & 0.876 & 0.823 \\
Conf $+$ $(1-\mathrm{sig})$ & 0.793 & 0.775 & 0.736 \\
Agree $+$ $(1-\mathrm{sig})$ & 0.819 & 0.835 & 0.786 \\
Full $R(x)$ & 0.833 & \textbf{0.879} & 0.811 \\
\midrule
\multicolumn{4}{l}{gpt-4o (SFT), acc@100 $=0.822$} \\
Conf-only (MaxProb) & 0.945 & 0.951 & 0.889 \\
Agree-only & 0.931 & 0.943 & 0.886 \\
$(1-\mathrm{sig})$-only & 0.797 & 0.798 & 0.765 \\
Conf $+$ Agree & 0.942 & 0.962 & 0.898 \\
Conf $+$ $(1-\mathrm{sig})$ & 0.875 & 0.882 & 0.828 \\
Agree $+$ $(1-\mathrm{sig})$ & 0.933 & 0.945 & 0.883 \\
Full $R(x)$ & 0.940 & \textbf{0.965} & 0.896 \\
\midrule
\multicolumn{4}{l}{gpt-4o (Standard DPO), acc@100 $=0.825$} \\
Conf-only (MaxProb) & 0.860 & 0.876 & 0.830 \\
Agree-only & 0.926 & 0.938 & 0.882 \\
$(1-\mathrm{sig})$-only & 0.800 & 0.803 & 0.754 \\
Conf $+$ Agree & 0.923 & \textbf{0.942} & 0.887 \\
Conf $+$ $(1-\mathrm{sig})$ & 0.831 & 0.835 & 0.791 \\
Agree $+$ $(1-\mathrm{sig})$ & 0.925 & 0.931 & 0.881 \\
Full $R(x)$ & 0.925 & 0.922 & 0.882 \\
\midrule
\multicolumn{4}{l}{gpt-4o (RA-DPO), acc@100 $=0.828$} \\
Conf-only (MaxProb) & 0.857 & 0.879 & 0.839 \\
Agree-only & 0.940 & 0.949 & 0.888 \\
$(1-\mathrm{sig})$-only & 0.807 & 0.818 & 0.770 \\
Conf $+$ Agree & 0.942 & 0.960 & 0.893 \\
Conf $+$ $(1-\mathrm{sig})$ & 0.848 & 0.853 & 0.810 \\
Agree $+$ $(1-\mathrm{sig})$ & 0.942 & 0.948 & 0.890 \\
Full $R(x)$ & 0.942 & \textbf{0.962} & 0.891 \\
\bottomrule
\end{tabular}
\caption{Inference-time ablation of $R(x)$. AUARC: area under the accuracy-rejection curve. Bold: best @50 per block; deltas in the text use unrounded values.}
\label{tab:rx_ablation}
\end{table}

\paragraph{Subset composition.}
Because agreement carries weight $0.8$ in $R_{\mathrm{train}}$, Smart-30\% contains only unanimous pairs, and larger subsets add progressively noisier 5/1, 4/2 and 3/3 labels (Appendix~\ref{app:ablation}, Table~\ref{tab:subset}). Composition therefore differs sharply across subsets while F1 stays near 0.82 from 30\% to 100\% of the data, as it does for a subset with no reliability structure at all.

Label quality still matters at the extreme, which the Ambiguous-only control confirms: training only on contested pairs drops gpt-4o to F1 = 0.653, below every other variant including the base. Its $\gamma$ weight rises to 0.207 (against 0.03--0.13 for the other gpt-4o variants), because agreement weakens as a predictor of correctness once the model is trained on noise, and, relative to the other DPO variants, the normalised weights shift toward the other two terms.

Splitting the test set by annotator agreement shows what the filter targets: on the 426 easy cases (agreement $\geq 0.83$) the fine-tuned variants other than Ambiguous-only reach $91$--$94\%$, while on the 266 hard cases (agreement $< 0.67$) the base reaches $61\%$ against their $63$--$67\%$ (Appendix~\ref{app:ablation}, Figure~\ref{fig:hard_cases}). Abstention removes exactly this band.

\begin{table*}[t]
\centering
\footnotesize
\setlength{\tabcolsep}{4pt}
\renewcommand{\arraystretch}{1.08}
\begin{tabular}{lcccccccccc}
\toprule
Variant & \multicolumn{6}{c}{EXIST} & \multicolumn{4}{c}{EDOS} \\
\cmidrule(lr){2-7}\cmidrule(lr){8-11}
 & \multicolumn{2}{c}{OpenAI} & \multicolumn{2}{c}{Qwen} & \multicolumn{2}{c}{Llama} & \multicolumn{2}{c}{Llama} & \multicolumn{2}{c}{Qwen} \\
\cmidrule(lr){2-3}\cmidrule(lr){4-5}\cmidrule(lr){6-7}\cmidrule(lr){8-9}\cmidrule(lr){10-11}
 & F1 & @50 & F1 & @50 & F1 & @50 & F1 & @50 & F1 & @50 \\
\midrule
SFT & 0.820 & \textbf{0.965} & \textbf{0.580} & 0.679 & 0.369 & 0.633 & 0.681 & \textbf{0.932} & 0.546 & 0.666 \\
Smart-10\% & 0.800 & 0.925 & 0.521 & 0.685 & 0.510 & 0.656 & -- & -- & -- & -- \\
Smart-30\% & 0.821 & 0.945 & 0.530 & 0.679 & 0.574 & 0.668 & 0.656 & 0.879 & 0.698 & 0.887 \\
Random-30\% & -- & -- & 0.544 & 0.679 & 0.575 & 0.673 & 0.648 & 0.904 & 0.692 & 0.880 \\
Smart-50\% & 0.819 & 0.939 & 0.550 & 0.662 & 0.594 & \textbf{0.711} & -- & -- & -- & -- \\
Random-50\% & 0.814 & 0.945 & 0.550 & 0.665 & 0.589 & 0.702 & 0.666 & 0.890 & 0.695 & 0.895 \\
Standard DPO & 0.821 & 0.922 & 0.566 & 0.691 & 0.596 & 0.697 & 0.681 & 0.831 & 0.711 & 0.913 \\
Ambiguous-only & 0.653 & 0.775 & 0.525 & 0.665 & 0.537 & 0.665 & -- & -- & -- & -- \\
RA-DPO & \textbf{0.826} & 0.962 & 0.572 & \textbf{0.697} & \textbf{0.600} & 0.676 & \textbf{0.689} & 0.835 & \textbf{0.725} & \textbf{0.920} \\
\bottomrule
\end{tabular}
\caption{F1 and accuracy at 50\% coverage across backbones and corpora (true agreement). Bold: column maximum.}
\label{tab:local}
\end{table*}

\paragraph{Generality across backbones and datasets.}
The framework runs unchanged on both corpora and on two open-weight 3B backbones, Qwen (Qwen2.5-3B-Instruct) and Llama (Llama-3.2-3B-Instruct). The evaluation prompt, pair ranking and out-of-fold $R(x)$ protocol are held fixed; the classifier, its fine-tuning (API or LoRA), the corpus and its token scorer change, and EDOS runs fewer variants (Table~\ref{tab:local}). Absolute numbers on the 3B backbones are lower because the bases are weaker (F1 $0.524$ for Qwen and $0.522$ for Llama against $0.724$ for OpenAI; Table~\ref{tab:threshold}); EDOS scores higher than EXIST on those same backbones, plausibly because its decision is English-only. The 3B backbones serve as corroboration rather than competition: they test whether the matched-budget result and the gain of reliability-gated abstention over full coverage reproduce on models whose weights are public and whose fine-tuning we control, not whether a 3B model matches the hosted one on absolute F1.

RA-DPO has the highest F1 on four of the five corpus-backbone tracks, forming a top cluster with SFT and Standard DPO on the fifth. Abstention raises accuracy over full coverage everywhere (Figure~\ref{fig:coverage} for EXIST; Tables~\ref{tab:edos} and~\ref{tab:edos_qwen25_3b} for EDOS), but which variant it favours at 50\% coverage varies: RA-DPO leads on Qwen for both corpora, SFT on gpt-4o and on EDOS-Llama, and Smart-50\% on EXIST-Llama. Ambiguous-only DPO is the weakest fine-tune on gpt-4o; on the 3B backbones the 554-pair Smart-10\% variant is lower still ($0.510$ and $0.521$ against $0.537$ and $0.525$). Two cells need explanation: SFT collapsed on EXIST-Llama (F1 $= 0.369$) and on EDOS-Qwen ($0.546$), an interaction between chat templates and our single-token YES/NO completion format; threshold-tuning restores $0.593$ and $0.704$ (Table~\ref{tab:threshold}), so both are calibration failures, not ranking failures, and every DPO variant on the same backbones recovers at the default threshold.

On EDOS the same findings hold and one sharpens (Tables~\ref{tab:edos} and~\ref{tab:edos_qwen25_3b}). Fine-tuning lifts F1 from $0.626$ to $0.689$ on Llama and from $0.666$ to $0.725$ on Qwen, and abstention raises every variant at 50\% coverage on both. Which variant leads is track-dependent: RA-DPO has the highest F1 on both EDOS backbones, but its margin over Standard DPO on Llama is not significant ($p = 0.33$), and on that backbone SFT is ahead on paired correctness ($p < 0.001$). EDOS's $n = 4{,}000$ split makes these distinctions possible where the EXIST intervals are twice as wide. The EDOS gains survive the threshold audit: RA-DPO exceeds the base's best-threshold ceiling ($0.689$ vs $0.645$; $0.725$ vs $0.681$) and keeps the highest F1 when every variant is tuned ($0.718$, $0.727$; Table~\ref{tab:threshold}). What transfers is the framework, not a ranking among variants.

The learned weights make the adaptation explicit (Table~\ref{tab:weights} for EXIST, Table~\ref{tab:edos} for EDOS). Matched on the same Llama backbone, every DPO variant shifts weight from agreement to confidence as the signal coarsens from four levels to two ($\beta$ falls from $0.535$ to $0.357$, from $0.572$ to $0.397$, and from $0.668$ to $0.426$ for Smart-30\%, Standard and RA-DPO, with $\alpha$ rising correspondingly). On EDOS-Llama, dropping the agreement term costs RA-DPO, the variant with the largest $\beta$, $4.7$\,pp at 50\% coverage and the base under half a point, yet confidence alone ($0.912$) exceeds both settings.

\section{Discussion}

Our results separate two findings that are easy to conflate. Training-side selection gives data efficiency: on gpt-4o, 30\% of the pairs matches full-data DPO, while on the 3B backbones it falls 1--4\,pp short. But a size-matched random draw (50\% on gpt-4o, 30\% on the 3B backbones) performs comparably on every corpus and backbone, so the ranking is not what produces the saving; a decision-threshold audit further separates gpt-4o's fine-tuning gains, which exceed any recalibration of its base, from the EXIST gains on the 3B backbones, which do not. Inference-side selection is where the score contributes most: on gpt-4o with true agreement, $R(x)$ lifts RA-DPO $8.4$\,pp above confidence alone at 50\% coverage ($0.9$\,pp with predicted agreement), and abstention raises every training variant on both corpora.

The learned $R(x)$ weights and the component ablation both show that annotator agreement is the most informative reliability signal for the DPO variants on gpt-4o. When the agreement signal coarsens, as on EDOS, confidence carries more weight, and on the weaker open-weight backbones it often becomes the strongest single abstention signal, consistent with \citet{varshney2022investigating}. The combined score does not recover that signal: on all four open-weight tracks confidence alone gives higher accuracy at 50\% coverage than $R(x)$ for RA-DPO and Standard DPO. The token-uncertainty factor contributes least at inference ($-0.3$\,pp when removed) and selects training pairs nearly at chance by itself; we keep it because on gpt-4o its weight rises when the model is trained only on contested pairs (the Ambiguous-only case above) and it alone is defined at token granularity.

The category analysis (Table~\ref{tab:category-analysis}) ties abstention to the task: at 50\% coverage RA-DPO abstains on $59$--$66\%$ of each sexist category and none of the 143 posts that carry no category (true-agreement setting): the gate routes posts that annotators dispute as sexist to human review.

On gpt-4o, the deployable gain with predicted agreement comes from a modest predictor (Pearson $r = 0.351$, the highest test-split $r$ among six regressor configurations of four encoders in Table~\ref{tab:regressor-comparison}), so better agreement prediction is the most direct way to narrow the gap to the true-agreement setting. Without annotator labels, SFT ranked by its own confidence ($0.951$) still exceeds every predicted-agreement score, and among the four variants of Table~\ref{tab:rx_ablation} predicted agreement adds over MaxProb only for RA-DPO.

\section{Conclusion}

RA-DPO is a reliability-aware selection and abstention framework on standard DPO, tested on three backbones and two corpora. On gpt-4o, 30\% of the pairs match full-data DPO, and on every track a size-matched random draw performs as well as the selected one: the gain is data efficiency, not selection quality; gpt-4o's gains, unlike the EXIST gains on the 3B backbones, exceed threshold recalibration. Abstention improves accuracy at matched coverage on every track, with the learned score strongest for RA-DPO on gpt-4o and confidence alone stronger for RA-DPO and Standard DPO on the open-weight backbones: on the hosted model, disagreement is worth more as an abstention signal than as a training filter.

\section*{Limitations}

\paragraph{Language scope.} Evaluation covers English and Spanish (EXIST 2023) and English (EDOS). The Structured prompt and the DPO recipe are language-agnostic in principle, but Appendix~\ref{app:openai_variants} shows a $3.4$\,pp English-Spanish gap on average across the base OpenAI versions, suggesting that prompt phrasing may need re-tuning for other languages. The agreement regressor is also markedly weaker on Spanish (Pearson $r = 0.231$) than on English ($r = 0.443$), which accounts for part of the gap between the predicted- and true-agreement settings and marks multilingual agreement prediction as a concrete direction for closing it.

\paragraph{Domain.} We test on short social-media posts. Longer-form text and other modalities are not evaluated; the token-uncertainty signal in $R(x)$ in particular is calibrated for short sequences.

\paragraph{Annotation protocols covered.} Our two corpora span six- and three-annotator protocols, which is the range currently available: sexism datasets that aggregate before release expose a single gold label and cannot fit $R(x)$'s $\beta$ term directly, so on those the deployable predicted-agreement variant is the applicable form of the method. Both corpora here are also social media text, labelled by crowdworkers on EXIST and by trained annotators with expert adjudication on EDOS; larger expert panels or other text domains may distribute agreement differently.

\paragraph{Combined score on weaker backbones.} On the open-weight backbones the learned combination in $R(x)$ ranks test items less well than model confidence alone for RA-DPO and Standard DPO, on both corpora. The out-of-fold weights are fit to predict correctness, not to maximise accuracy at a given coverage, and with a weaker classifier that fit does not recover the confidence ranking; on such a backbone, $R(x)$ should be compared against MaxProb on held-out data before it is adopted.

\paragraph{Base-model scope on the OpenAI track.} Every DPO variant trains on a single hosted base, so per-base generalisation within the OpenAI family is unverified; the open-weight track covers two further backbones.

 \paragraph{Threshold $\tau$.} $\tau$ is chosen by maximising the harmonic mean of accuracy and coverage over a threshold sweep. The value is task-
  and model-specific; deployment to a new setting would require re-tuning $\tau$ on held-out data. A sensitivity analysis (Appendix~\ref{app:ablation}, Table~\ref{tab:tau_sensitivity}) shows the choice is not fragile: accuracy moves by up to about two points for $\tau$ within $\pm 0.05$ of the selected value, and in a split-half check where $\tau$ is tuned on one half of the data and applied to the other, unseen-half accuracy matches the full-data value (RA-DPO on gpt-4o: mean $0.835$, standard deviation $0.015$ across 200 random half-splits of the 692 test items).

 \section*{Ethical Considerations}

\paragraph{Responsible inference.}
A core design choice of RA-DPO is to abstain when the reliability score $R(x)$ falls below a threshold $\tau$. Rather than assigning a label to contested or ambiguous content, the model declines to answer, which acts as a built-in safeguard against overconfident outputs on subjective inputs. We recommend that any deployment combine this abstention rule with human review for high-stakes decisions.

\paragraph{Data and sensitive content.}
We use EXIST 2023 \citep{plaza2023exist} and EDOS \citep{kirk2023edos}, two benchmarks released for research on sexism detection. Both contain material that targets women and is, by design, offensive; we added a content warning at the top of the paper, kept dataset examples out of the figures and tables, and constructed a synthetic illustrative tweet for Figure~\ref{fig:pipeline} so that no real instance from either dataset is reproduced verbatim in the paper.

\paragraph{Openness and reproducibility.}
Two of the three backbones we evaluate (the Llama and Qwen 3B models) are open-weight and run on a single accelerator. We release code, prompts, hyperparameters and per-instance prediction arrays for every fine-tuned variant on GitHub (link below the abstract), with an archived snapshot on Zenodo,\footnote{\url{https://doi.org/10.5281/zenodo.21802837}} so the main findings can be audited without depending on closed APIs. The dataset texts are not redistributed. The arrays reproduce the true-agreement coverage-accuracy results and the threshold sensitivity analysis; the other results also need the official dataset releases or, for the hosted model, API access.

\section*{Acknowledgments}

We thank the three anonymous reviewers, whose reports led to the component ablation, the second corpus, and the additional baselines in this version, and the area chair for the meta-review. We also thank the organisers and annotators of EXIST 2023 and EDOS for releasing the individual annotator labels this work builds on. We acknowledge support from the focus area Applied Data Science (ADS) at Utrecht University.

\bibliography{references}

\appendix

\renewcommand{\topfraction}{0.98}
\renewcommand{\bottomfraction}{0.98}
\renewcommand{\dbltopfraction}{0.98}
\renewcommand{\textfraction}{0.02}
\renewcommand{\floatpagefraction}{0.95}
\renewcommand{\dblfloatpagefraction}{0.95}
\setcounter{topnumber}{4}
\setcounter{dbltopnumber}{3}
\setcounter{bottomnumber}{3}
\setcounter{totalnumber}{8}

\section{OpenAI Model Variants}
\label{app:openai_variants}

\begin{figure}[htbp]
\centering
\includegraphics[width=\columnwidth]{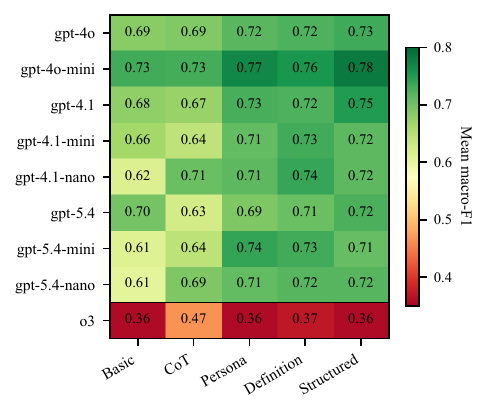}
\caption{Mean macro-F1 over zero-/few-shot and both languages, per OpenAI version and prompt strategy.}
\label{fig:heatmap}
\end{figure}

The main paper reports results on three backbones (OpenAI, Llama, Qwen) and uses one OpenAI base for all fine-tuning. Seven further OpenAI versions \citep{openai2024gpt4omini, openai2025gpt41, openai2026gpt54, openai2026gpt54mini} and a reasoning model \citep[o3;][]{openai2025o3} set the prompt-strategy baselines in zero-shot and few-shot. Table~\ref{tab:models} reports the best F1 per version over prompts, shot settings and language subsets; Figure~\ref{fig:heatmap} gives the mean F1 per version and prompt strategy.

\begin{table}[htbp]
\centering
\fontsize{9}{12}\selectfont
\setlength{\tabcolsep}{6pt}
\begin{tabular}{lcc}
\toprule
Model & F1 & Best setting \\
\midrule

gpt-4o            & 0.803 & Persona/FS/EN \\
gpt-4o-mini       & 0.792 & Struct./FS/ES \\
gpt-4.1           & 0.772 & Persona/FS/EN \\
gpt-4.1-mini      & 0.777 & Persona/FS/EN \\
gpt-4.1-nano      & 0.762 & Def./ZS/EN \\
gpt-5.4           & 0.746 & Struct./FS/EN \\
gpt-5.4-mini      & 0.768 & Def./FS/EN \\
gpt-5.4-nano      & 0.749 & Struct./ZS/ES \\
o3                & 0.573 & CoT/FS/EN \\

\bottomrule
\end{tabular}
\caption{Best macro-F1 per OpenAI version over prompts, shot settings and language subsets (no fine-tuning). ZS/FS: zero-/few-shot; Def./Struct.: Definition/Structured; EN/ES: English/Spanish.}
\label{tab:models}
\end{table}

\begin{table}[htbp]
\centering
\small
\begin{tabular}{lcc}
\toprule
Strategy & Avg F1 & Std \\
\midrule
Structured & 0.731 & 0.031 \\
Definition & 0.728 & 0.028 \\
Persona & 0.722 & 0.040 \\
Chain-of-Thought & 0.675 & 0.041 \\
Basic & 0.661 & 0.080 \\
\bottomrule
\end{tabular}
\caption{Average macro-F1 by prompt strategy (8 base OpenAI versions; the
o3 reasoning model and the fine-tuned model are excluded).}
\label{tab:prompts}
\end{table}

The o3 reasoning model performs poorly (best F1 $0.573$; $0.36$--$0.47$ when averaged per prompt). On this binary task extended reasoning does not help, and the extra generation step appears to add noise to the YES/NO answer.

Averaged over prompts and shot settings, F1 is higher on English than on Spanish for seven of the eight base OpenAI versions, by $2.6$ to $6.7$\,pp for six of them, and the gap averages $3.4$\,pp across all eight; gpt-4o-mini ($-0.6$\,pp) and gpt-5.4-mini ($+0.6$\,pp) perform about the same in both languages.

\section{Hyperparameters and Compute}
\label{app:hyperparams}

\paragraph{OpenAI fine-tuning.} SFT and every DPO variant on the gpt-4o base use the OpenAI fine-tuning API with its automatic batch size, learning-rate multiplier and epoch count (resolved to batch size 2, multiplier 1.0 and 2 epochs for Smart-30\%), and the DPO variants use $\beta = 0.1$; the three single-component retrains of Table~\ref{tab:training-ablation} set these values explicitly to match. All variants draw from the same pool of 5{,}536 majority-vote preference pairs (opposite label as rejected); the Smart-$k\%$ and control variants differ only in subset selection, and RA-DPO submits all 5{,}536 pairs with agreement-based duplication (two copies when agreement $\geq 5/6$; 8{,}984 rows).

\paragraph{Open-weight fine-tuning.} Both Llama-3.2-3B-Instruct and Qwen2.5-3B-Instruct use low-rank adaptation \citep[LoRA;][]{hu2022lora} with $r=16$, $\alpha=32$, dropout $0.05$, on the q/k/v/o projection modules. SFT: 1 epoch, learning rate $2 \times 10^{-5}$, per-device batch size 2 with gradient accumulation 8. DPO: 1 epoch, $\beta = 0.1$, learning rate $5 \times 10^{-6}$, per-device batch size 1 with gradient accumulation 16, max prompt length 384, max sequence length 512. Inference: max new tokens 8, fp32 throughout.

\paragraph{$R(x)$ weight learning.} Per-fold logistic regression on the three-feature vector $(\mathrm{conf}, \mathrm{agree}, 1-\mathrm{sig})$ predicting correctness. 5-fold stratified cross-validation, seed 42. Coefficients are normalised so $\alpha + \beta + \gamma = 1$.

\paragraph{Agreement regressor.} Twitter-XLM-RoBERTa-base (\texttt{cardiffnlp/\allowbreak twitter-\allowbreak xlm-\allowbreak roberta-\allowbreak base}; \citealp{barbieri2022xlmt}) fine-tuned as a regressor on the training split predicting $\mathrm{agree}(x) \in [0.5, 1]$ from text alone. Learning rate $3 \times 10^{-5}$, batch size 32, 5 epochs, weight decay $0.01$, warmup ratio $0.1$, early stopping patience 2 on validation mean squared error. Test set: Pearson $r = 0.351$, Spearman $\rho = 0.328$, mean absolute error (MAE) $= 0.135$. Per language, $r = 0.443$ on English against $0.231$ on Spanish (MAE $0.132$ vs $0.139$). This model was selected from six regressor configurations (four encoders, plus two training-schedule variants of the adopted one), ranked by test-split $r$ (Table~\ref{tab:regressor-comparison}): it improves on mBERT ($r = 0.308$; \citealp{devlin2019bert}) and vanilla XLM-R base ($r = 0.322$; \citealp{conneau2020unsupervised}), while XLM-R large underfits on this data size ($r = 0.128$).

\begin{table}[htbp]
\centering
\footnotesize
\setlength{\tabcolsep}{2pt}
\resizebox{\columnwidth}{!}{%
\begin{tabular}{lccc}
\toprule
Encoder & Pearson $r$ & Spearman $\rho$ & MAE \\
\midrule
Twitter-XLM-R base$^\dagger$      & 0.351 & 0.328 & 0.135 \\
Twitter-XLM-R base (7 ep.)        & 0.342 & 0.338 & 0.134 \\
XLM-R base                        & 0.322 & 0.300 & 0.138 \\
Twitter-XLM-R (Spearman stop)  & 0.310 & 0.335 & 0.138 \\
mBERT                             & 0.308 & 0.276 & 0.139 \\
XLM-R large                       & 0.128 & 0.118 & 0.146 \\
\bottomrule
\end{tabular}%
}
\caption{Agreement-regressor encoder comparison on the test split ($n = 692$). $\dagger$ marks the adopted model.}
\label{tab:regressor-comparison}
\end{table}

\paragraph{Abstention threshold $\tau$.} Selected by sweeping $\tau \in [0.30, 0.95]$ on a 50-point grid and picking the value that maximises the harmonic mean of accuracy and coverage (Eq.~\ref{eq:harmonic}); the split-half check in Table~\ref{tab:tau_sensitivity} shows that accuracy at the selected threshold transfers across halves of the data.

\paragraph{Compute.} OpenAI experiments run through the API; most of the cost is the eleven fine-tuning jobs (eight variants plus three single-component retrains). Open-weight experiments run on a single 64\,GB accelerator in fp32; one DPO variant takes 1--2 hours to fine-tune and a few minutes to run inference over the 692-sample test set.

\paragraph{EDOS setup.} EDOS \citep{kirk2023edos} releases the individual labels of all three annotators per item, so the reliability score applies without modification. We use the official splits, the adjudicated gold label as ground truth, the same Structured prompt at evaluation, the same $R_{\mathrm{train}}$ pair ranking and reliability-weighted duplication, and the same out-of-fold $R(x)$ protocol; token-uncertainty scores are computed once with Llama-3.2-3B-Instruct's token probabilities and shared by both backbones. We run the seven variants that the budget comparison needs (base, SFT, Standard DPO, Smart-30\%, Random-30\%, Random-50\%, and RA-DPO), so that every EDOS selection arm has a size-matched random counterpart.

\section{Per-Variant Results, Ablations and Robustness}
\label{app:ablation}

This appendix isolates each $R(x)$ component and probes the robustness of abstention.

\begin{figure}[htbp]
\centering
\includegraphics[width=\columnwidth]{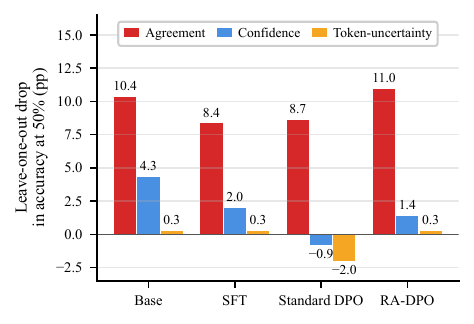}
\caption{Leave-one-out importance of each $R(x)$ component on gpt-4o: accuracy drop at 50\% coverage when it is removed.}
\label{fig:loo}
\end{figure}

\paragraph{Inference-time ablation.} Table~\ref{tab:rx_ablation} in Section~\ref{sec:results} ranks test items by each component alone, by each pair, and by the full score. Ranking by confidence alone is the MaxProb baseline \citep{varshney2022investigating}. Because agreement takes only four values, coverage cuts can land inside a tie group; we therefore report the expected accuracy under uniform random tie-breaking for every rule. Two-feature and three-feature rules use out-of-fold logistic weights (5-fold stratified cross-validation), so no rule ever sees the item it scores. Figure~\ref{fig:loo} summarises the resulting leave-one-out importance: removing agreement costs $8$--$11$\,pp at 50\% coverage on all four variants shown (and up to $13.6$\,pp on the remaining gpt-4o selection variants), whereas, once agreement is present, token-level uncertainty adds at most $0.3$\,pp and confidence $1.4$--$4.3$\,pp on the base, SFT and RA-DPO (none on Standard DPO). On Standard DPO, where the confidence$+$agreement rule is already near-optimal, restoring the remaining components lowers the full score by $2.0$\,pp, a gap that does not reach significance under a paired bootstrap over test items ($95\%$ CI $[0.0, 3.2]$, $p = 0.07$). Full-coverage accuracy is identical across rules within a block and appears in each block header. On the open-weight backbones the picture differs: at 50\% coverage, confidence alone outranks the full score for RA-DPO and Standard DPO on both corpora (Section~\ref{sec:results}).

\paragraph{Per-variant results in all three agreement settings.} Table~\ref{tab:coverage-complete} extends the coverage analysis of Table~\ref{tab:coverage} to every gpt-4o variant, and Tables~\ref{tab:edos} and~\ref{tab:edos_qwen25_3b} give the same grid for both EDOS backbones with bootstrap 95\% CIs ($2{,}000$ resamples, $n = 4{,}000$). On gpt-4o, in the deployable predicted-agreement setting, SFT (0.890 at 50\% coverage) and RA-DPO (0.887) are statistically indistinguishable. RA-DPO improves on its own no-agreement baseline by $3.5$\,pp and on its confidence alone by $0.9$; SFT's confidence alone already reaches $0.951$ (Table~\ref{tab:rx_ablation}), above its predicted-agreement $0.890$, so among the four variants of Table~\ref{tab:rx_ablation} predicted agreement adds over MaxProb only for RA-DPO.

On EDOS-Llama, fine-tuning raises F1 from $0.626$ $[0.611, 0.641]$ at base to between $0.648$ and $0.689$ across variants, and Smart-30\% reaches higher accuracy than full-data DPO ($0.755$ vs $0.729$) on $30\%$ of the pairs, though its F1 is $2.5$\,pp lower: on this corpus the two metrics move in opposite directions, a precision/recall trade-off on the minority class. RA-DPO has the highest F1 ($0.689$ $[0.675, 0.705]$), though its margin over Standard DPO is not significant (McNemar $p = 0.33$), and SFT is the stronger model on paired correctness ($p < 0.001$). At a matched $30\%$ budget Smart-30\% and Random-30\% are statistically tied ($0.656$ vs $0.648$, $p = 0.22$), the EDOS counterpart of the EXIST result in Section~\ref{sec:results}; the other three 3B matched-budget comparisons are not significant either (McNemar $p = 1.00$ on EXIST-Llama, $0.58$ on EXIST-Qwen and $1.00$ on EDOS-Qwen; F1 in Table~\ref{tab:local}).

A text-predicted agreement score is competitive with the true one on EDOS. The same regressor encoder, trained on the EDOS train split, reaches $r = 0.30$ against a binary target, and substituting its predictions for the annotator labels matches or exceeds the true-agreement setting on most variants (Llama Standard DPO, $0.831$ to $0.856$ at 50\% coverage). Both settings stay above the no-agreement floor for every variant. With only two agreement levels the true label is a coarse ranking feature, so a continuous estimate from text loses little and sometimes orders the test set better.

\paragraph{Learned weights across corpora.} Section~\ref{sec:results} quantifies how the weight shifts from agreement to confidence. Two details belong here. First, $\alpha$ rises as $\beta$ falls, from $0.067$ to $0.564$ (Smart-30\%), $0.240$ to $0.455$ (Standard) and $0.193$ to $0.378$ (RA-DPO); on Smart-30\% the rise exceeds the fall in $\beta$ because the token-uncertainty weight also drops ($0.398$ to $0.08$), so the score redistributes weight across the other two terms rather than losing it. Second, we exclude SFT from that comparison: its EXIST counterpart on the same backbone is the collapsed run discussed in Section~\ref{sec:results}, so its weights are not a valid reference point.

\begin{table}[htbp]
\centering
\small
\setlength{\tabcolsep}{4pt}
\begin{tabular}{llccc}
\toprule
Method & Setting & @100 & @60 & @50 \\
\midrule
Base         & True      & 0.740 & 0.834 & 0.879 \\
             & Pred.     & 0.740 & 0.824 & 0.835 \\
             & No-agr.   & 0.740 & 0.793 & 0.775 \\
\addlinespace
SFT          & True      & 0.822 & 0.940 & 0.965 \\
             & Pred.     & 0.822 & 0.894 & 0.890 \\
             & No-agr.   & 0.822 & 0.875 & 0.882 \\
\addlinespace
Smart-10\%     & True      & 0.802 & 0.911 & 0.925 \\
             & Pred.     & 0.802 & 0.872 & 0.884 \\
             & No-agr.   & 0.802 & 0.805 & 0.801 \\
\addlinespace
Smart-30\%     & True      & 0.824 & 0.937 & 0.945 \\
             & Pred.     & 0.824 & 0.855 & 0.853 \\
             & No-agr.   & 0.824 & 0.836 & 0.824 \\
\addlinespace
Smart-50\%     & True      & 0.821 & 0.937 & 0.939 \\
             & Pred.     & 0.821 & 0.870 & 0.882 \\
             & No-agr.   & 0.821 & 0.831 & 0.827 \\
\addlinespace
Random-50\%    & True      & 0.816 & 0.935 & 0.945 \\
             & Pred.     & 0.816 & 0.868 & 0.887 \\
             & No-agr.   & 0.816 & 0.819 & 0.809 \\
\addlinespace
Standard DPO & True      & 0.825 & 0.925 & 0.922 \\
             & Pred.     & 0.825 & 0.858 & 0.867 \\
             & No-agr.   & 0.825 & 0.831 & 0.835 \\
\addlinespace
Ambiguous-only    & True      & 0.697 & 0.764 & 0.775 \\
             & Pred.     & 0.697 & 0.745 & 0.786 \\
             & No-agr.   & 0.697 & 0.701 & 0.691 \\
\addlinespace
RA-DPO       & True      & 0.828 & 0.942 & 0.962 \\
             & Pred.     & 0.828 & 0.872 & 0.887 \\
             & No-agr.   & 0.828 & 0.848 & 0.853 \\
\bottomrule
\end{tabular}
\caption{Accuracy at matched coverage under three agreement settings, for every gpt-4o variant.}
\label{tab:coverage-complete}
\end{table}

\begin{table*}[t]
\centering
\footnotesize
\setlength{\tabcolsep}{3pt}
\begin{tabular}{lrcccccc}
\toprule
Variant & Pairs & F1 (95\% CI) & Acc@100 & Acc@50 & Acc@50 (Pred.) & Acc@50 (no-agr.) & $\alpha/\beta/\gamma$ \\
\midrule
Base (prompted) & -- & 0.626 [0.611, 0.641] & 0.661 & 0.779 & 0.781 & 0.775 & 0.62/0.27/0.11 \\
SFT & 14,000 & 0.681 [0.664, 0.698] & 0.787 & 0.932 & 0.927 & 0.923 & 0.69/0.27/0.04 \\
Standard DPO & 14,000 & 0.681 [0.667, 0.696] & 0.729 & 0.831 & 0.856 & 0.813 & 0.45/0.40/0.15 \\
Smart-30\% & 4,200 & 0.656 [0.639, 0.673] & 0.755 & 0.879 & 0.902 & 0.840 & 0.56/0.36/0.08 \\
Random-30\% & 4,200 & 0.648 [0.631, 0.666] & 0.759 & 0.904 & 0.911 & 0.867 & 0.56/0.38/0.05 \\
Random-50\% & 7,000 & 0.666 [0.650, 0.683] & 0.749 & 0.890 & 0.903 & 0.862 & 0.55/0.35/0.10 \\
RA-DPO & 14,000 & 0.689 [0.675, 0.705] & 0.724 & 0.835 & 0.854 & 0.788 & 0.38/0.43/0.20 \\
\bottomrule
\end{tabular}
\caption{EDOS results on Llama-3.2-3B-Instruct ($n = 4{,}000$). Pred.\ uses text-predicted agreement; no-agr.\ drops it; $\alpha/\beta/\gamma$: learned weights.}
\label{tab:edos}
\end{table*}

\begin{table*}[t]
\centering
\footnotesize
\setlength{\tabcolsep}{3pt}
\begin{tabular}{lrcccccc}
\toprule
Variant & Pairs & F1 (95\% CI) & Acc@100 & Acc@50 & Acc@50 (Pred.) & Acc@50 (no-agr.) & $\alpha/\beta/\gamma$ \\
\midrule
Base (prompted) & -- & 0.666 [0.651, 0.682] & 0.708 & 0.856 & 0.845 & 0.837 & 0.61/0.34/0.05 \\
SFT & 14,000 & 0.546 [0.532, 0.562] & 0.550 & 0.666 & 0.668 & 0.659 & 0.64/0.27/0.09 \\
Standard DPO & 14,000 & 0.711 [0.695, 0.727] & 0.774 & 0.913 & 0.892 & 0.872 & 0.50/0.43/0.08 \\
Smart-30\% & 4,200 & 0.698 [0.681, 0.714] & 0.772 & 0.887 & 0.891 & 0.850 & 0.48/0.44/0.08 \\
Random-30\% & 4,200 & 0.692 [0.675, 0.708] & 0.772 & 0.880 & 0.893 & 0.859 & 0.51/0.44/0.05 \\
Random-50\% & 7,000 & 0.695 [0.679, 0.711] & 0.773 & 0.895 & 0.895 & 0.845 & 0.50/0.42/0.08 \\
RA-DPO & 14,000 & 0.725 [0.709, 0.740] & 0.786 & 0.920 & 0.906 & 0.884 & 0.52/0.45/0.03 \\
\bottomrule
\end{tabular}
\caption{EDOS results on Qwen2.5-3B-Instruct ($n = 4{,}000$). Pred.\ uses text-predicted agreement; no-agr.\ drops it; $\alpha/\beta/\gamma$: learned weights.}
\label{tab:edos_qwen25_3b}
\end{table*}

\begin{figure}[htbp]
\centering
\includegraphics[width=\columnwidth]{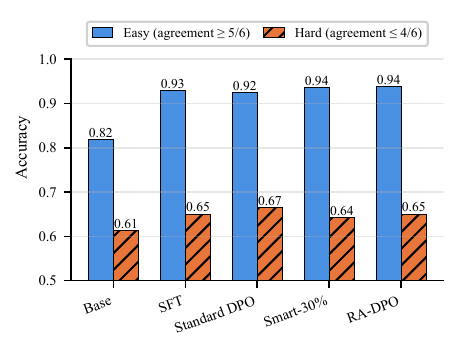}
\caption{Accuracy on easy vs.\ hard test cases for the gpt-4o base and four fine-tuned variants.}
\label{fig:hard_cases}
\end{figure}

\paragraph{Hard vs.\ easy cases.} Figure~\ref{fig:hard_cases} splits the test set by annotator agreement: 266 hard cases (agreement $< 0.67$) against 426 easy ones ($\geq 0.83$). Every DPO variant of Table~\ref{tab:finetuning} except the Ambiguous-only control reaches at least $0.908$ on the easy split (Figure~\ref{fig:hard_cases} plots three of them); the spread on the hard split is what abstention targets.

\paragraph{Subset composition by agreement level.} Table~\ref{tab:subset} gives the agreement-level composition of every training subset: 3/3 split (agreement $=0.5$), 4/2 ($0.667$), 5/1 ($0.833$) and 6/0 unanimous ($1.0$). Smart-30\% is entirely unanimous; Smart-50\% is 65\% unanimous with the remainder at 5/1; Random-50\% and Standard both carry roughly 12\% fully contested 3/3 pairs. The same fixed formula behaves differently on EDOS, where three annotators give two agreement levels: in $R_{\mathrm{train}}(x) = 0.8 \cdot \text{agree} + 0.2 \cdot (1 - \text{sig})$ a disputed item scores at most $0.733$ and a unanimous one at least $0.800$, so the blocks cannot overlap and Smart-30\% there is drawn entirely from the unanimous pool, the top $4{,}200$ of its $10{,}850$ items under the token-uncertainty tie-break. The duplication rule covers all $10{,}850$ of them.

\begin{table}[htbp]
\centering
\fontsize{10}{12}\selectfont
\setlength{\tabcolsep}{3pt}
\begin{tabular}{lccccc}
\toprule
Subset & Pairs & 3/3 & 4/2 & 5/1 & 6/0 \\
\midrule

Smart-10\%   & 554   & 0\%   & 0\%  & 0\%  & 100\% \\
Smart-30\%   & 1,661 & 0\%   & 0\%  & 0\%  & 100\% \\
Smart-50\%   & 2,768 & 0\%   & 0\%  & 35\% & 65\% \\
Random-50\%  & 2,768 & 12\%  & 26\% & 30\% & 32\% \\
Standard DPO & 5,536 & 12\%  & 26\% & 30\% & 33\% \\
Ambiguous-only  & 665   & 100\% & 0\%  & 0\%  & 0\% \\

\bottomrule
\end{tabular}
\caption{Composition of DPO subsets by annotator agreement.}
\label{tab:subset}
\end{table}

\begin{table*}[t]
\centering
\footnotesize
\setlength{\tabcolsep}{4pt}
\begin{tabular}{lcccc}
\toprule
 & \multicolumn{2}{c}{Llama} & \multicolumn{2}{c}{Qwen} \\
\cmidrule(lr){2-3}\cmidrule(lr){4-5}
Selection & F1 & @50 & F1 & @50 \\
\midrule
Random, whole pool & 0.543 [0.502, 0.579] & 0.673 & 0.533 [0.494, 0.570] & 0.679 \\
Reliability, stratified & 0.550 [0.510, 0.587] & 0.679 & 0.530 [0.492, 0.568] & 0.673 \\
Drop contested stratum & 0.533 [0.495, 0.570] & 0.656 & 0.529 [0.490, 0.566] & 0.665 \\
\midrule
Random, unanimous only & 0.512 [0.475, 0.550] & 0.671 & 0.528 [0.490, 0.565] & 0.679 \\
Smart-10\% (original rule) & 0.510 [0.472, 0.547] & 0.656 & 0.521 [0.482, 0.558] & 0.685 \\
Highest uncertainty & 0.477 [0.441, 0.511] & 0.656 & 0.511 [0.472, 0.547] & 0.671 \\
\midrule
Wrong-first, agreement gated & 0.582 [0.544, 0.618] & 0.691 & 0.575 [0.536, 0.613] & 0.685 \\
Label-matched random, gated & 0.588 [0.549, 0.626] & 0.691 & 0.575 [0.536, 0.613] & 0.697 \\
Wrong-first, ungated & 0.582 [0.543, 0.618] & 0.691 & 0.575 [0.536, 0.613] & 0.702 \\
\midrule
Full data (5,536) & 0.596 & 0.697 & 0.566 & 0.691 \\
Tuned threshold, no training & 0.604 & -- & 0.587 & -- \\
\bottomrule
\end{tabular}
\caption{Selection at a matched 554-pair budget: distribution-matched, unanimous-tail, and error-ranked blocks; last row: tuned base threshold, no training. F1 with 95\% CIs.}
\label{tab:lowbudget}
\end{table*}

\paragraph{Training losses.} On Qwen the DPO training losses order as RA-DPO ($0.499$, lowest) $<$ Smart-50\% ($0.514$) $<$ Smart-30\% ($0.545$) $<$ Standard ($0.563$) $<$ Random-50\% ($0.599$) $<$ Smart-10\% ($0.638$) $<$ Ambiguous-only ($0.693$); Llama reproduces the ordering within $\pm 0.01$. As the matched random controls show, a lower training loss on a cleaner subset does not translate into higher test F1.

\paragraph{Selection at a low budget.} Table~\ref{tab:lowbudget} isolates why reliability selection underperforms at 554 pairs, the budget with the most headroom left. Nine arms (variants trained at the same 554-pair budget) fall into three blocks; the error-ranked block is discussed two paragraphs below. All $p$-values in this paragraph are McNemar tests over the 692 paired test items. In the unanimous-tail block the selection narrows the training set: the original Smart-10\% rule costs $3.3$\,pp against a random draw from the pool on Llama ($p = 0.02$), a random draw from the same tail costs $3.0$\,pp ($p = 0.03$), and the 554 most uncertain unanimous pairs cost $6.5$\,pp ($p = 0.02$). In the distribution-matched block the pool's agreement distribution is preserved and the penalty disappears: ranking by $R_{\mathrm{train}}$ within each agreement stratum recovers $4.0$\,pp on Llama and lands level with random ($+0.7$\,pp, $p = 0.25$; $-0.2$\,pp on Qwen).

Two things follow. The cost of reliability selection here is distribution narrowing rather than the reliability signal itself: holding the agreement distribution fixed removes the entire penalty. But the signal does not add anything either, since the stratified arm only reaches parity with random rather than exceeding it. Label noise does not explain the gap: dropping the fully contested stratum, whose pairs are so unreliable that training on them alone drops gpt-4o to $0.653$ from a prompted $0.724$, does not improve on a random draw that keeps them ($-1.0$\,pp Llama, $-0.4$\,pp Qwen). Injecting noise directly gives the same result: relabelling every pair with one randomly drawn annotator's vote flips $17\%$ of the random arm's labels and none of the all-unanimous selection's, yet costs the random arm only $1.7$\,pp on Llama and $0.4$\,pp on Qwen, so it still matches the selection ($p = 0.18$ and $p = 0.69$). Within a stratum the token-uncertainty score cannot discriminate in any case: its standard deviation across unanimous pairs is $0.013$.

\begin{table*}[t]
\centering
\footnotesize
\setlength{\tabcolsep}{4pt}
\begin{tabular}{lcccccccccc}
\toprule
 & \multicolumn{2}{c}{EXIST-OpenAI} & \multicolumn{2}{c}{EXIST-Llama} & \multicolumn{2}{c}{EXIST-Qwen} & \multicolumn{2}{c}{EDOS-Llama} & \multicolumn{2}{c}{EDOS-Qwen} \\
\cmidrule(lr){2-3}\cmidrule(lr){4-5}\cmidrule(lr){6-7}\cmidrule(lr){8-9}\cmidrule(lr){10-11}
Model & default & tuned & default & tuned & default & tuned & default & tuned & default & tuned \\
\midrule
Base (prompted) & 0.724 & 0.764 & 0.522 & 0.606 & 0.524 & 0.595 & 0.626 & 0.645 & 0.666 & 0.681 \\
SFT & 0.820 & 0.828 & 0.369 & 0.593 & 0.580 & 0.600 & 0.681 & 0.690 & 0.546 & 0.704 \\
Standard DPO & 0.821 & 0.827 & 0.596 & 0.602 & 0.566 & 0.593 & 0.681 & 0.689 & 0.711 & 0.713 \\
RA-DPO & 0.826 & 0.831 & 0.600 & 0.613 & 0.572 & 0.599 & 0.689 & 0.718 & 0.725 & 0.727 \\
\midrule
Base, tuned on validation & -- & -- & -- & 0.604 & -- & 0.587 & -- & -- & -- & -- \\
\bottomrule
\end{tabular}
\caption{Decision-threshold audit: F1 at the default threshold and at the best swept threshold. Last row tunes the base threshold on validation.}
\label{tab:threshold}
\end{table*}

\begin{table*}[t]
\centering
\small
\setlength{\tabcolsep}{3pt}
\begin{tabular}{lcccccc}
\toprule
Model & $\tau^*$ & Acc@$\tau^*$ & Cov@$\tau^*$ & Acc range $\pm 0.05$ & Stable width & Split-half acc \\
\midrule
Base (gpt-4o) & 0.300 & 0.740 & 1.000 & 0.740--0.740 & 0.358 & 0.740 $\pm$ 0.018 \\
SFT (gpt-4o) & 0.300 & 0.822 & 1.000 & 0.822--0.822 & 0.305 & 0.821 $\pm$ 0.015 \\
Standard DPO (gpt-4o) & 0.300 & 0.825 & 1.000 & 0.825--0.825 & 0.212 & 0.825 $\pm$ 0.014 \\
RA-DPO (gpt-4o) & 0.526 & 0.835 & 0.991 & 0.828--0.849 & 0.252 & 0.835 $\pm$ 0.015 \\
RA-DPO (Llama) & 0.592 & 0.636 & 0.984 & 0.629--0.650 & 0.292 & 0.633 $\pm$ 0.019 \\
RA-DPO (Qwen) & 0.552 & 0.636 & 0.997 & 0.634--0.644 & 0.345 & 0.635 $\pm$ 0.018 \\
\bottomrule
\end{tabular}
\caption{Sensitivity of the abstention threshold $\tau$, with a 200-run split-half stability check (mean $\pm$ SD).}
\label{tab:tau_sensitivity}
\end{table*}

\paragraph{Ranking by base-model error.} A revised rule, the error-ranked block of Table~\ref{tab:lowbudget}, follows reducible-loss selection \citep{mindermann2022prioritized}: keep pairs whose label is trustworthy (agreement $\geq 5/6$) and rank them by how confidently the base model contradicts the label. At 554 pairs this arm reaches $0.582$ on Llama and $0.575$ on Qwen, above the random draw by $4.0$ and $4.2$\,pp (paired bootstrap $p < 0.001$). Its own controls show the gain does not come from the error signal: both base models place essentially all confident errors on YES-labelled items, a label-matched random draw from the same gated pool scores the same ($0.588$ and $0.575$), and removing the gate changes nothing. The error signal is informative at the class level, where it identifies which label the model under-predicts, and uninformative at the item level, like every other per-item signal in the table.

\paragraph{Threshold audit.} Table~\ref{tab:threshold} reports F1 at the default threshold and at the best swept threshold for every track. A decision threshold tuned on the validation split and applied to the untrained base reaches $0.604$ on Llama and $0.587$ on Qwen, above every 554-pair arm and above full-data DPO on both backbones, so at this scale the low-budget gains, including the class-composition effect above, are recalibrations of the operating point. The audit separates the tracks: every fine-tuned OpenAI variant except the Ambiguous-only control exceeds the base's best-threshold ceiling ($0.800$--$0.826$ against $0.764$; default F1s in Table~\ref{tab:finetuning}), RA-DPO exceeds it on both EDOS backbones and keeps the highest tuned F1 there ($0.718$, $0.727$), and the two SFT collapses recover under tuning (from $0.369$ to $0.593$ and from $0.546$ to $0.704$), a calibration failure, not a ranking one. Selective-prediction comparisons are unaffected, since $R(x)$ ranks within one model.

\paragraph{Training-selection overlap.} Table~\ref{tab:selection-overlap} quantifies how much of the Smart-30\% subset is determined by agreement alone. The train split contains 1,805 unanimous pairs competing for 1,661 slots, so any agreement-only selection overlaps Smart-30\% by at least 91.3\% regardless of tie-breaking. An uncertainty-only selection overlaps at chance level and its composition matches the agreement distribution of the train split.

\begin{table}[htbp]
\centering
\footnotesize
\setlength{\tabcolsep}{2pt}
\resizebox{\columnwidth}{!}{%
\begin{tabular}{lccccc}
\toprule
Selection & Overlap & 3/3 & 4/2 & 5/1 & 6/0 \\
\midrule
Smart-30\% ($R_{\mathrm{train}}$) &      100.0\% & 0.0\%  & 0.0\%  & 0.0\%  & 100.0\% \\
Agree-only top-30\%               &  91.3--100\% & 0.0\%  & 0.0\%  & 0.0\%  & 100.0\% \\
Conf-only top-30\%                &       48.5\% & 4.9\%  & 12.2\% & 27.6\% & 55.2\% \\
Uncert-only top-30\%              &       28.4\% & 12.8\% & 28.1\% & 30.8\% & 28.4\% \\
Random 30\% (expected)            &       30.0\% & 12.0\% & 25.7\% & 29.7\% & 32.6\% \\
\bottomrule
\end{tabular}%
}
\caption{Overlap of alternative 1,661-pair training selections with Smart-30\%.}
\label{tab:selection-overlap}
\end{table}

\paragraph{Training-side ablation.} Table~\ref{tab:training-ablation} completes the picture by retraining DPO on top-30\% selections ranked by a single component each, on all three backbones. On gpt-4o the ordering is informative: the $R_{\mathrm{train}}$ ranking (Smart-30\%, F1 $0.821$) is statistically tied with uncertainty-only ($0.814$, McNemar $p = 0.44$) and significantly better than agreement-only ($0.797$, $p = 0.03$). We report that gap without reading it as evidence for the token-uncertainty term, since within the unanimous block that score varies by only $0.013$ and the two selections differ mostly in which unanimous pairs the tie-break admits. On the weaker 3B backbones all agreement- and uncertainty-based selections land in one statistically indistinguishable band (Llama $0.56$--$0.58$, $p \geq 0.34$; Qwen $0.53$--$0.54$), and the agreement-only rows bound the tie-break noise floor ($\Delta$F1 $\leq 0.006$). The consistent exception is confidence-only selection, which keeps the 1{,}661 pairs the prompted gpt-4o base is most confident about (the same pairs on all three backbones) and fails on every backbone: on gpt-4o it is indistinguishable from the prompted base ($0.731$ vs $0.724$, $p = 0.75$), and on the 3B models it collapses ($0.386$ Llama, $0.466$ Qwen). The cause is class skew, not pair difficulty: the gpt-4o base is near-certain almost exclusively on non-sexist posts, so the confidence-only top-30\% contains just $10.8\%$ positive (sexist) pairs. Its selection consists mostly of unanimous, overwhelmingly non-sexist items, as the agreement-level composition in Table~\ref{tab:selection-overlap} confirms, and DPO on that skew reduces minority-class recall. External agreement does not have this failure mode, which is a concrete argument for reliability signals beyond model confidence in selection.

\begin{table}[htbp]
\centering
\footnotesize
\setlength{\tabcolsep}{3pt}
\resizebox{\columnwidth}{!}{%
\begin{tabular}{lcccccc}
\toprule
 & \multicolumn{2}{c}{gpt-4o} & \multicolumn{2}{c}{Llama} & \multicolumn{2}{c}{Qwen} \\
\cmidrule(lr){2-3}\cmidrule(lr){4-5}\cmidrule(lr){6-7}
Selection (top 30\%) & F1 & @50 & F1 & @50 & F1 & @50 \\
\midrule
Smart-30\% ($R_{\mathrm{train}}$) & 0.821 & 0.945 & 0.574 & 0.668 & 0.530 & 0.679 \\
Agreement-only & 0.797 & 0.919 & 0.561 & 0.691 & 0.526 & 0.668 \\
Agreement-only (alt.\ tie-break) & -- & -- & 0.567 & 0.665 & 0.527 & 0.691 \\
Uncertainty-only & 0.814 & 0.948 & 0.583 & 0.682 & 0.544 & 0.679 \\
Confidence-only & 0.731 & 0.879 & 0.386 & 0.613 & 0.466 & 0.633 \\
\bottomrule
\end{tabular}%
}
\caption{Training-side ablation: DPO on the top 30\% of pairs (1,661) ranked by one component, against Smart-30\%.}
\label{tab:training-ablation}
\end{table}

\paragraph{Agreement-weighted and soft-label SFT baselines.} To test whether simpler alternatives to preference training suffice, we add two SFT baselines that use the same annotator signal: agreement-weighted SFT, whose per-example loss is scaled by agreement, and soft-label SFT, which trains toward the annotator distribution \citep{wu-etal-2023-dont}. Both require per-example loss weights or soft targets, which the hosted OpenAI fine-tuning API does not support, so they run on the open-weight track (Table~\ref{tab:reweighted-sft}). The outcome is mixed: on Llama, DPO variants stay ahead (F1 $0.596$--$0.600$ vs $0.560$--$0.578$), and agreement-weighted SFT recovers most of the plain-SFT template collapse; on Qwen, both baselines are statistically tied with the DPO variants. Agreement-weighted SFT is therefore a reasonable baseline at full coverage, while the preference-pair interface remains what unifies selection, weighting, and abstention in one score.

\begin{table}[htbp]
\centering
\footnotesize
\setlength{\tabcolsep}{2pt}
\begin{tabular}{lcccc}
\toprule
 & \multicolumn{2}{c}{Llama} & \multicolumn{2}{c}{Qwen} \\
\cmidrule(lr){2-3}\cmidrule(lr){4-5}
Method & F1 & @50 & F1 & @50 \\
\midrule
SFT (plain) & 0.369 & 0.633 & 0.580 & 0.679 \\
Agreement-weighted SFT & 0.578 & 0.630 & 0.585 & 0.694 \\
SFT, soft labels & 0.560 & 0.604 & 0.583 & 0.694 \\
Standard DPO & 0.596 & 0.697 & 0.566 & 0.691 \\
RA-DPO & 0.600 & 0.676 & 0.572 & 0.697 \\
\bottomrule
\end{tabular}
\caption{Agreement-weighted and soft-label SFT baselines on the open-weight backbones.}
\label{tab:reweighted-sft}
\end{table}

\begin{figure}[htbp]
\centering
\includegraphics[width=\columnwidth]{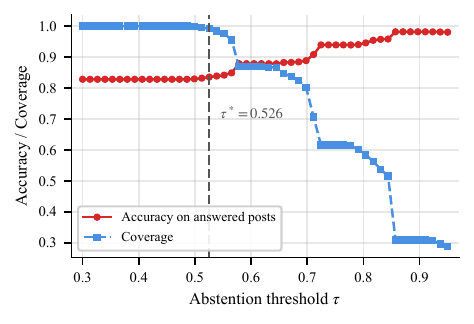}
\caption{Accuracy and coverage vs.\ the abstention threshold $\tau$ for RA-DPO (gpt-4o); $\tau^*$ marked.}
\label{fig:tau}
\end{figure}

\paragraph{Threshold sensitivity.} Table~\ref{tab:tau_sensitivity} reports how accuracy and coverage respond to the abstention threshold. $\tau^*$ maximises the harmonic mean of accuracy and coverage on the test sweep, and the stable interval is the widest contiguous range whose accuracy stays within 1\,pp of accuracy at $\tau^*$. Around each model's $\tau^*$, accuracy moves by up to about two points within $\pm 0.05$; in a 200-run split-half experiment ($\tau$ tuned on one random half of the test items, evaluated on the other), unseen-half accuracy matches the full-data value for every model, and for RA-DPO on gpt-4o the selected threshold itself is stable across splits (mean $0.524$, standard deviation $0.033$, against $0.526$ on the full test set); on the 3B backbones it varies more (standard deviation $0.13$ and $0.10$) while unseen-half accuracy still matches. For Base, SFT and Standard DPO the optimal $\tau^* = 0.30$ leaves coverage at $1.0$, so their bands are flat; RA-DPO is the case where abstention is active. Figure~\ref{fig:tau} plots the full curve for RA-DPO.

\paragraph{Abstention by sexism category.} Table~\ref{tab:category-analysis} breaks RA-DPO's abstention behaviour down by the most frequent EXIST Task 3 category among annotators' votes, in the true-agreement setting; a post enters a category if any annotator assigns one, so 239 of the 549 categorised posts carry a not-sexist majority-vote label (58 of them split 3/3). The gate concentrates on posts whose sexism annotators dispute: it abstains on none of the 143 posts without a sexism category and on 59--66\% of each sexist category, raising retained accuracy to 0.89--1.00 per category. Grouped instead by the plurality EXIST Task 2 source-intention label, the gate abstains on $61\%$ of direct, $60\%$ of judgemental and $49\%$ of reported posts, against $34\%$ of posts with no plurality intention label, and retained accuracy is $0.94$--$0.98$ in every group.

\begin{table}[htbp]
\centering
\footnotesize
\setlength{\tabcolsep}{2pt}
\resizebox{\columnwidth}{!}{%
\begin{tabular}{lrccc}
\toprule
Category & $n$ & Acc@100 & Acc@50 & Abstain\% \\
\midrule
No sexism category & 143 & 0.986 & 0.986 & 0.0 \\
Ideological inequality & 194 & 0.747 & 0.944 & 63.4 \\
Misogyny (non-sexual viol.) & 79 & 0.810 & 0.893 & 64.6 \\
Objectification & 105 & 0.829 & 1.000 & 65.7 \\
Sexual violence & 58 & 0.776 & 0.909 & 62.1 \\
Stereotyping \& dominance & 113 & 0.805 & 0.957 & 59.3 \\
\bottomrule
\end{tabular}%
}
\caption{RA-DPO (gpt-4o, true agreement) per sexism category: accuracy at full and 50\% coverage, and the share of each category abstained on.}
\label{tab:category-analysis}
\end{table}

\end{document}